\documentclass[sigconf]{acmart}

\definecolor{ForestGreen}{RGB}{34,139,34}
\usepackage{color,soul}
\usepackage{subcaption}
\usepackage{pdflscape}
\usepackage{colortbl}
\usepackage{placeins}
\AtBeginDocument{%
  \providecommand\BibTeX{{%
    \normalfont B\kern-0.5em{\scshape i\kern-0.25em b}\kern-0.8em\TeX}}}

\copyrightyear{2025}
\acmYear{2025}
\setcopyright{acmlicensed}\acmConference[CHI '25]{CHI Conference on Human Factors in Computing Systems}{April 26-May 1, 2025}{Yokohama, Japan}
\acmBooktitle{CHI Conference on Human Factors in Computing Systems (CHI '25), April 26-May 1, 2025, Yokohama, Japan}
\acmDOI{10.1145/3706598.3714009}
\acmISBN{979-8-4007-1394-1/25/04}

\acmSubmissionID{7187}

\begin{document}

\title{The Robotability Score: Enabling Harmonious Robot Navigation\\on Urban Streets}

\author{Matt Franchi}
\authornote{The authors contributed equally to this research.}
\email{mattfranchi@cs.cornell.edu}
\affiliation{
    \institution{Cornell University, Cornell Tech}
    \state{New York}
    \country{USA}
}
\author{Maria Teresa Parreira}
\authornotemark[1]
\email{mb2554@cornell.edu}
\affiliation{
    \institution{Cornell University, Cornell Tech}
    \state{New York}
    \country{USA}
}
\author{Fanjun Bu}
\authornotemark[1]
\email{fb266@cornell.edu}
\affiliation{
    \institution{Cornell University, Cornell Tech}
    \state{New York}
    \country{USA}
}

\author{Wendy Ju}
\email{wendyju@cornell.edu}
\affiliation{
    \institution{Jacobs Technion-Cornell Institute, Cornell Tech}
    \state{New York}
    \country{USA}
}

\renewcommand{\shortauthors}{Franchi, Parreira \& Bu et al.}

\begin{abstract}
This paper introduces the Robotability Score ($R$), a novel metric that quantifies the suitability of urban environments for autonomous robot navigation. Through expert interviews and surveys, we identify and weigh key features contributing to R for wheeled robots on urban streets. Our findings reveal that pedestrian density, crowd dynamics and pedestrian flow are the most critical factors, collectively accounting for 28\% of the total score. Computing robotability across New York City yields significant variation; the area of highest R is 3.0 times more ``robotable'' than the area of lowest R. Deployments of a physical robot on high and low robotability areas show the adequacy of the score in anticipating the ease of robot navigation. This new framework for evaluating urban landscapes aims to reduce uncertainty in robot deployment while respecting established mobility patterns and urban planning principles, contributing to the discourse on harmonious human-robot environments.

\end{abstract}



\begin{teaserfigure}
    \centering
  \includegraphics[width=0.9\textwidth]{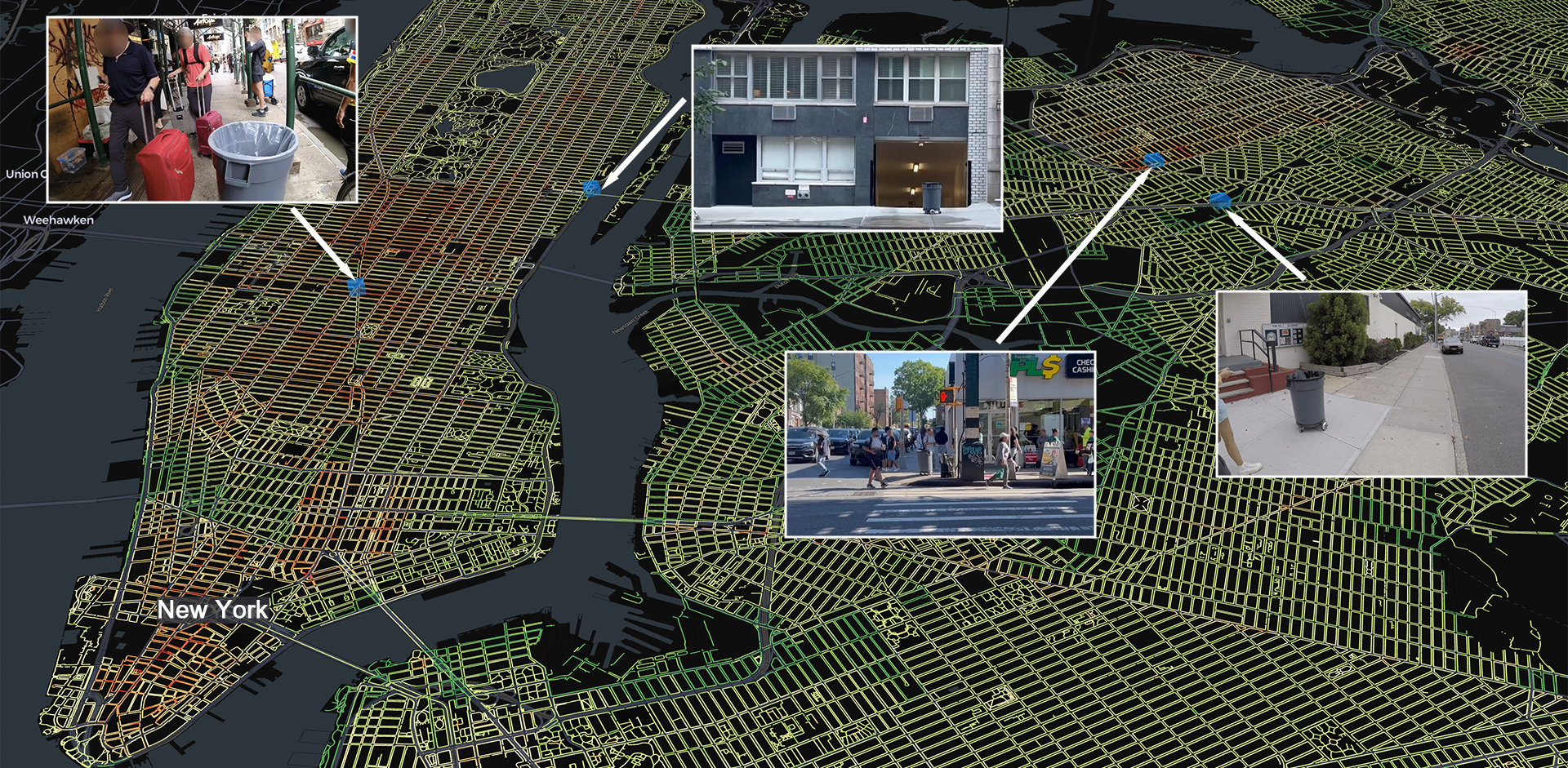}
  \caption{Robotability-informed mobile trashcan robot deployments in New York City, September 2024. An interactive web application describing robotability is available at \href{https://robotability.cornell.edu/}{robotability.cornell.edu}.}
  \Description{Robotability-informed mobile trashcan robot deployments in New York City, September 2024.}
  \label{fig:teaser}
\end{teaserfigure}

\received{20 February 2007}
\received[revised]{12 March 2009}
\received[accepted]{5 June 2009}

\maketitle

\newcommand{\numExpertInterviews}{10\space}
\newcommand{\numSurveyResponses}{40\space}
\newcommand{\numSidewalkPoints}{1.9 million \space}

\section{Introduction}
\label{sec:intro}

The urban landscape is one of fixed space, where emerging stakeholders compete for navigable area against longstanding transportation mediums and components of the built environment. In the early 20th century,  cities changed drastically as personal vehicles and public transportation came to replace carriages \cite{chiu_evolution_2008}. Now, micromobility platforms like e-scooters and e-bikes are exploding in popularity; cities are adding bike lanes en force, often via the redistribution of existing vehicular parking \cite{wild_beyond_2018} or traffic lanes \cite{barnes_improving_2013}.  %

In the ever-changing urban fabric, emerging literature suggests a real potential for robots to become prevalent actors (for example, autonomous sidewalk delivery robots \cite{hossain_autonomous_2023, Grush2022}).
As robotic systems become increasingly visible on public streets, there is a growing need for methods to evaluate the suitability of urban landscapes for robot navigation \cite{mavrogiannis2023challenges, tiddi2019rci}. This work introduces the \textbf{Robotability Score ($R$)}, a novel metric designed to measure how the built environment and its occupancy patterns support robot navigation (moving from point A to B) in public urban spaces.
Robotability considers a range of features that impact robot navigation, including physical attributes of the space (e.g., sidewalk width), urban activity flows (e.g., pedestrian density) and functional considerations specific to robotic systems (e.g., connectivity), and combines these features into a single score given feature importance to enable seamless navigation. By providing a systematic approach to evaluate urban landscapes for robot navigability, the Robotability Score aims to reduce uncertainty about how and where to deploy mobile robots in public, shared spaces. It is crucial to note that the purpose of this score is not to inform robot-first city planning over human needs. Instead, it aims to generate an awareness of how robotic systems may interact with existing urban designs, working towards a more harmonious integration of these technologies without compromising the primacy of human-centric urban planning principles.

The Robotability Score can thus be formulated as a tool for robot navigation, anticipatory urban design, and interaction modeling, distilling expert knowledge and available datasets into a useful, evaluative human-robot interaction metric. The concept of robotability is generalizable and adapts to custom robot architectures with adjustments to the features and weights included in the calculation of the score. In this study, we focus on \textit{wheeled sidewalk robots} in a prototyping of $R$. To determine a set of input features that influence robot navigation, we conduct \numExpertInterviews expert interviews, followed by an online survey where experts are asked to rank these features through pairwise comparisons. From this, we extract a set of weights that allow us to consolidate the features into a single index, the Robotability Score. Finally, we validate robotability through four targeted proof-of-concept deployments of a wheeled trashcan robot \cite{bu_trash_2023} in various neighborhoods of New York City (NYC).

Towards reproducibility and customizability, we publish our study materials and analysis code. The code to reproduce a proof-of-concept computation of $R$ in New York City, survey materials, and code to observe our computation of $R$'s feature weights, is available at \href{https://robotability.cornell.edu/}{robotability.cornell.edu}.

\section{Related Work} 
\label{sec:relwork}

\subsection{Urban sensing}
Cities are learning to utilize the wealth of data that they produce and procure to analyze aspects of their functionality with unparalleled granularity. An epitomic example of this is pedestrian volume counts; historically collected on a sampling of intersections by manual surveyors, tools like cell phone mobility data have come about that map movement trends across entire cities \cite{nilforoshan_human_2023}. %
Other sources of data are emerging as ``sensors'' for urban flows, such as panoramic street view imagery (\cite{anguelov_google_2010, naik_computer_2017, cinnamon_-it-yourself_2022}), large-scale dashcam data \cite{franchi_detecting_2023, shapira_fingerprinting_2024, franchi_towards_2024, chowdhury_tracking_2021}, and on-edge privacy-preserving cameras for the monitoring of traffic \cite{noauthor_numina_nodate}.

In this data landscape, urban planners have an opportunity to enact \textit{anticipatory} changes in their cities through vision making \cite{poli_vision_2017}, modeling \cite{gober_urban_2016} and data analysis \cite{maffei_data-driven_2020}; for example, a prominent problem being tackled by these methods is urban climate adaptation \cite{ndatabaye_anticipatory_2024, birchall_anticipatory_2021, gober_urban_2016}. With our proposed Robotability Score, we introduce a forward-thinking trend of anticipating changes in the sidewalk ballet \cite{kasinitz_metropolis_1995}, a tool that could facilitate harmony come the day that robots arrive on city streets. 

\subsection{Scoring the urban environment}

The Robotability Score inherits from a rich literature of designing and generating metrics to measure aspects of interactions with urban environments and perform inter-city and infra-city comparisons \cite{mahajan_global_2024, boeing_urban_2019, lowry_comparing_2014, conderino_social_2021,krambeck_global_2006}. Biazzo et al. \cite{biazzo_citychrone_2022, biazzo_general_2019} developed a ``velocity score'' to represent average traveling times between locations, and a ``sociality score'' to represent the average number of possible encounters on public transit in a typical day. %
The JANE index \cite{delclos-alio_looking_2018}, an attempt to operationalize the beliefs of renowned urbanist Jane Jacobs, was developed from 22 variables that are aggregated to model urban vitality \cite{gomez-varo_jane_2022}. More static parts of the urban fabric, like street greenery, have also been measured \cite{li_assessing_2015}. The Walk Score \footnote{\url{https://www.walkscore.com}} and its analogous Transit Score and Bike Score (also studied more broadly under ``Bikeability'' \cite{silva_bike_2011}) weigh a set of features from the environment to produce single-dimensional data that informs users on what to expect from different spaces. The Walk Score, for example, considers factors such as population density, distance to amenities, and block length, compiling these into point systems and a final score between 0 and 100. Other scores that support route-planning for accessibility and particular mobility needs include AccessMap \cite{bolten2019access} and Project Sidewalk \cite{saha2019sidewalk}.%

Analogously to these scores, robotability is conceptualized as an index that describes urban spaces from the perspective of mobile robots, following recent literature which rethinks Robot-City Interactions \cite{tiddi2019rci}. Wheeled robot navigation shares commonalities with wheelchair navigation with respect to physical constraints when navigating a public urban space. Prior work on assessing accessibility \cite{ferreira2007index, bolten2019access, mumm_accessibility_2024} indicates factors like sidewalk slope, volume of vehicle traffic, sidewalk condition, climate/weather conditions, sidewalk width, and curb ramp accessibility. A Robotability Score must consider, in addition to these aspects of mobility, constraints that pertain to networking and online navigation (e.g. connectivity infrastructure, 5G or WiFi access) or power (e.g. charging stations) \cite{liu2020energy}.

\subsection{Robot deployments in urban streets}

As robots continue to develop autonomy, their presence on public streets is becoming more common, drawing attention from both academia and industry. Many researchers are envisioning a future where robots share sidewalks with people, exploring the potential interactions between passersby and robots in real-world settings \cite{TrashInMotion, OutOfPlaceRobotIntheWild, FromAgentAutonomyToCausualCollab, UnderstandingInteraction, pelikan2024encountering}. Beyond interpersonal interactions, there is a growing focus on the relationship between robots and the urban environment. The concepts of ``smart streets'' and ``smart cities'' have prompted researchers to investigate opportunities and needs at an urban scale, towards the future cohabitation of all users \cite{tiddi2020robot, plank2022ready, sevtsuk2019future, while2021urban, lynn2020smart, puig2017public}. %

However, deploying robots in public streets poses significant challenges. Beyond logistical hurdles, the complex and dynamic nature of urban environments makes social navigation difficult \cite{mavrogiannis2023challenges, karwowski_quantitative_2023} and raises concerns about safety, privacy, and equity \cite{while2021urban}. Researchers have also highlighted the need for new policies to regulate the use of robots in public spaces \cite{mintrom2022robots, thomasen2020robots, salvini2010investigation, salvini2010investigation, woo2020urban, salvini2021safety, kovacic2023regulating, salvini2022safety, marks2019robots}.
Understanding the harmonious integration of Public-area Mobile Robots (PMRs) \cite{grush2024intelligent} onto the urban fabric thus requires organized collaboration. For example, the Knight Autonomous Vehicle Initiative brings together the University of Oregon with the cities of Detroit, Pittsburgh, San José, and Miami to pilot personal delivery devices in public areas \cite{howell2022piloting, weinberg2023sharing}.

While these collaborations provide valuable insights into the practical challenges of deploying PMRs, there remains a need for metrics to assess urban environments' suitability for robot navigation. Such metrics would not only facilitate these collaborative efforts but also provide a common evaluation for stakeholders across different cities and projects. In response to this need, we propose the Robotability Score.
To our best knowledge, this is the first work on developing a score which aggregates features from urban environments into a single indicator of the suitability of robot navigation. 

\section{Methods}

In this section, we detail the three-part methodology of our project: expert interviews and surveys, computing the Robotability Score $R$, and real-world validative robot deployments.  All materials are made available in the \href{https://robotability.cornell.edu/}{study repository}.
\label{sec:methods}

\subsection{Expert Interviews}
We ran a series of expert interviews to understand the needs, applications, and features (or indicators) of the Robotability Score. We looked for experts in the fields of Robotics, Accessibility, and Urban Planning, both in academia and industry. Recruitment was carried out through email invites. Experts were chosen based on the relevance of their work and snowball sampling (interviewee suggestions and contacts). Data was collected under Cornell University exempt IRB protocol IRB0148634. Interview participants provided informed consent. The online interview had a duration of 45 minutes; the interview script can be consulted in the study repository. We began by asking participants about their professional background and how it pertains to robot navigation. We introduced the concept of the Robotability Score and asked for their thoughts on which features may impact robotability. We then showed participants a list of features prepared \textit{a priori} and asked for their thoughts on the list. Finally, participants were asked to fill out a survey where they were shown pairwise combinations of those indicators/features and had to choose \textit{which feature} (of the pair) \textit{is more important for robot navigation in public spaces}, sharing their thoughts aloud. This approach allows us to pilot the survey, which is conceptualized as a method to collect expert insights more efficiently. Interview transcripts and notes were analyzed to extract insights and inform our survey design and feature set. 

\subsection{Survey for Pairwise Comparison}

To obtain $R$ for a specific configuration, environmental and functional features must be weighted according to their relative importance to the robot's navigation. The initial set of features was selected based on the authors' expertise and general consensus in literature on robot navigability. To obtain the set of weights for these features, we relied on expert insights through a methodology adapted from the Analytic Hierarchy Process (AHP) \cite{subramanian2012ahp, dede2021safety}, where rankers are shown pairwise combinations of items and assign their relative importance. This is combined to extract a set of weights as is detailed in Section \ref{sec:derive-weights} below. The full survey is designed using Qualtrics experience management software and can be found in the study repository. Participants are asked about their professional occupations. Then, they are introduced to the concept of the Robotability Score and the finalized list of features. For a set of indicator features $F$, we can obtain pairwise combinations, denoted by $C$. Participants ranked 50 randomly selected pairs of features (around 20\% of C for $|F|=24$)  by choosing the feature more important for robot navigation. This is a deviation from the AHP method, as experts usually rate all the possible combinations on a scale, which was deemed to cause cognitive load and loss of attention when piloting. The display order of each feature in a pair is also randomized. The survey takes around 20 minutes to complete.

Recruitment was performed through direct invites, professional social media, and snowball sampling. Experts interviewed were also asked to fill out the full survey.

\subsection{Derivation of Weights}
\label{sec:derive-weights}
To calculate the relative importance of the set of features \( F \) for the Robotability Score, we utilize pairwise comparisons and the eigenvalues of a contingency matrix. This approach ensures that the weights assigned to each feature are grounded in expert preferences, as expressed through pairwise rankings.

\subsubsection{Feature Set and Pairwise Combinations}
Let \( F = \{f_1, f_2, \dots, f_n\} \) represent a set of features under consideration. The set of all possible pairwise combinations of these features is denoted by \( C \), defined as:

\begin{equation}
    C = \{(f_i, f_j) \mid f_i, f_j \in F, \, i < j\}
\end{equation}

The total number of pairwise combinations is given by:
$|C| = \frac{n!}{(n-2)!2!}$

\subsubsection{Pairwise Rankings and Contingency Matrix}

Following the AHP methodology, for each pair \( (f_i, f_j) \in C \) there is a number of expert preference answers $C_{ij}$.  From this, we can build a contingency matrix \( M \). This matrix \( M \) is an \( n \times n \) matrix where each element \( M_{ij} \) represents the preference ratio for feature \( f_i \) over feature \( f_j \):

\begin{equation}
M_{ij} = \frac{f_i \text{ selected}} {f_j \text{ selected}}
\end{equation}

Note that \( M_{ij} = \frac{1}{M_{ji}} \) and \( M_{ii} = 1 \).

\subsubsection{Eigenvalue Decomposition for Weight Calculation}
To determine the weights of each feature, we perform eigenvalue decomposition on the matrix \( M \). The eigenvalues \( \lambda_1, \lambda_2, \dots, \lambda_n \) and corresponding eigenvectors \( \mathbf{v}_1, \mathbf{v}_2, \dots, \mathbf{v}_n \) are computed as follows:

\begin{equation}
M \mathbf{v}_k = \lambda_k \mathbf{v}_k \quad \text{for } k = 1, 2, \dots, n
\end{equation}

The weight \( w_i \) for each feature \( f_i \) is derived from the eigenvector \( \mathbf{v}_1 \) corresponding to the largest eigenvalue \( \lambda_1 \). Specifically, the weight \( w_i \) is calculated as:

\begin{equation}
w_i = \frac{v_{1i}}{\sum_{k=1}^{n} v_{1k}}
\end{equation}

The normalization ensures that the weights sum to one, providing a meaningful measure of the relative importance of each feature based on the pairwise comparisons.

Note that recalculating the weights based on unavailable sets of features requires merely removing those features from the contingency matrix and recalculating eigenvalues. Because of the normalization step, the customized weight set will still sum up to 1.

\subsection{Computing the Robotability Score, $R$}

Having computed the weight $w$ for each feature in the set of relevant features $F$, we design a formula for $R$, the Robotability Score. At a high level, we evaluate $R$ at every point in a spatial network (or, more technically, "node" $n$ in a graph structure), denoted as $G$. 

At each point, $R$ is influenced by the processed value of each feature at that point, the polarity of the feature (whether or not the feature adds or subtracts from the score), and the method by which features are combined. We combine features summatively. In processing the raw feature values, we apply normalization, or constraining the range of values in a feature's numerical distribution to lie between 0 and 1. 

Formally, the score \( R \) at a point is calculated using the following formula:

\begin{equation}
R = \sum_{i=1}^{|F|} p_i \cdot w_i \cdot x_i
\end{equation}

\( x_i \) represents the value of feature \( f_i \), and \( p_i \) represents the polarity of that feature. \( w_i \) and \( p_i \) determine the magnitude and direction of the contribution to the score \( R \). We determine the polarity for each feature $f_i$ via expert judgment, as parsed from the aforementioned interview transcripts.

\subsubsection{Feature Normalization and Interpretation}
For consistency, all features in $F$ need to be in a uniform distribution; we choose $[0,1]$ for simplicity. For features that are binary, no change is needed; for features with variable distributions, we apply min-max normalization \cite{moreira_effects_2021}. We note that while min-max normalization is a common and established method for feature normalization, there may be room for downstream work on more representative scaling methods \cite{mazziotta_normalization_2022}.

\subsubsection{Summation with Polarity Control}

 It is necessary to account for the fact that not all features contribute positively to $R$. For example, per expert consensus, an increase in pedestrian density would be detrimental to $R$, while an increase in proximity to charging stations would be beneficial. To address this, we propose a scoring model that includes a polarity control component, allowing each feature to either increase or decrease the aggregate score \( R \).

Given a set of features \( F = \{f_1, f_2, \dots, f_n\} \), and corresponding weights \( W = \{w_1, w_2, \dots, w_n\} \), we introduce a set of polarity indicators \( P = \{p_1, p_2, \dots, p_n\} \). Each polarity indicator \( p_i \) is defined as follows:

\begin{equation}
p_i = 
\begin{cases} 
+1 & \text{if } f_i \text{ contributes positively to the score } R, \\
-1 & \text{if } f_i \text{ contributes negatively to the score } R.
\end{cases}
\end{equation}

Having specified robotability from a theoretical perspective, we now move to define how $R$ might extend to a real setting: city sidewalks.

\subsubsection{Extending $R$ to a Sidewalk Graph \( G \)}
A natural (and intended) application of $R$ is to apply the score across an area with a real-world sidewalk network. We encapsulate the sidewalks of New York City through a graph $G$, which is a data structure inherently suited to spatial data as composed of nodes and edges, such as urban networks \cite{barthelemy_modeling_2008}. $R$ can be uniquely computed at the edge or node level of $G$, abstracting each length of the sidewalk. 

An additional processing step in our proof-of-concept implementation is that we \textit{segmentize} the sidewalk network of NYC. Segmentization means that we sample points along each sidewalk in $G$, at a customizable threshold (we choose $T$=15\text{m})\footnote{We select a segmentation threshold of 15m (approximately 50 feet) to balance computational complexity and the spatial precision of R.} \cite{franchi_estimating_2024}. Functionally, this means we can evaluate $R$ at several points along each sidewalk, which is useful for the more accurate capture of hyper-local features like street furniture density and pedestrian density. Before segmentization, we have 476 thousand thousand unique sidewalk edges; after, we have 1.9 million unique points of computation, a 4-fold increase in granularity. \\

We formalize the sidewalk network representation as follows. Let \( G = (N, E) \) be a road graph, where:
\begin{itemize}
    \item \( N \) represents the set of nodes (e.g., street intersections or segmentized along-sidewalk points), and
    \item \( E \) represents the set of edges (e.g., street segments or $T$-length connections between segmentized points) connecting these nodes.
\end{itemize}

Each node \( n \in N \) is associated with a set of features \( F_n = \{f_{n1}, f_{n2}, \dots, f_{nn}\} \) that describe relevant characteristics (such as traffic density, pavement quality, or speed limits) of the road along that node.

For each node \( n \in N \), we compute a score \( R_n \) using summation with polarity control, as defined previously.
Let \( x_{ni} \) represent the value of feature \( f_{ni} \) on node \( n \).

The score \( R_n \) for node \( n \) is then calculated as:

\begin{equation}
R_n = \sum_{i=1}^{|F|} p_{i} \cdot w_{i} \cdot x_{ni}
\end{equation}

\subsubsection{Aggregate Score for the Graph}
To obtain an aggregate score \( R_G \) for the entire graph \( G \), we compute the mean of the individual node scores \( R_n \). This aggregate score is given by:

\begin{equation}
R_G = \frac{1}{|N|} \sum_{n \in N} R_n
\end{equation}

This aggregate score \( R_G \) provides a measure of an entire place's robotability. In practice, it makes more sense to aggregate over some sub-grouping of $G$, like neighborhoods or districts; an example of this is aggregating over a semantically-meaningful geographic grouping like Census blocks (described in \ref{sec:Deployments}), which we use to determine the optimal locations for our proof-of-concept deployment.

\subsubsection{Feature extraction}
From our finalized list of features $F$, we undergo a data sourcing and processing task to represent each indicator. We take advantage of NYC's nationally-leading public data availability (available at NYC OpenData\footnote{https://opendata.cityofnewyork.us/}) to source data for many features \cite{cukier_rise_2013}; in fact, this is a major motivation behind choosing NYC as our location for proof-of-concept deployment. 

A full list of features computed for NYC and data sources can be found at \autoref{tab:indicators}. We use NYC OpenData datasets to represent 10 out of 24 indicators; for brevity here, we elaborate further in Appendix \ref{sec:feature-computation}. \label{sec:dashcam} Three traffic-related indicators are computed in-house using large-scale dashcam imagery collected from instrumented ridesharing vehicles \cite{franchi_towards_2024}.  Our data is sourced from Nexar, Inc. \cite{nexar_inc_nexar_2024}. Between August 11th, 2023, and August 31st, 2023, we collect a set of 7.6 million images via software we develop to download, preprocess, and verify fresh data. The set of images, $D$, is paired with geographic and temporal metadata that allow precise localization. For each dashcam image in our large set, we run an object detection model (YOLOv7 \cite{wang_yolov7_2022}) to assess the number of pedestrians, vehicles, and cyclists\footnote{We implement an intersection-over-union filter to prevent the case where a cyclist may also be counted as a pedestrian.}. The per-image object counts are then joined to the nearest node $n$ in $N$, and averaged to measure the average traffic experienced at $n$. These average counts then map directly to vehicle traffic, pedestrian density, and bicycle traffic. We note that this measure is not intended to be comprehensive or extremely precise; our performance threshold is to see \textit{differences} across different levels of the city \footnote{The statistical validity of this approach is supported by the near-citywide coverage of our distribution, and prior work on fingerprinting foot traffic with dashcam data \cite{franchi_towards_2024}. The dashcam images are distributed throughout all 5 boroughs of NYC; within our segmentized distribution of 1.9 million points, 330 thousand, or 17.4\%, are missing data. When aggregating by Census blocks, only 0.4\% of blocks are missing image data coverage; the median Census block contains 62 images.}
. One can consider our computed traffic counts as a lower bound on the real, unknown traffic distribution (or more generally, the true distribution of an object of interest \cite{franchi_detecting_2023}).

We model certain features, including weather conditions, with a uniform value across the entire graph $G$. This assumption applies when the feature to be represented is expected or measured to have no variation at the time of score application (e.g., at deployment time ) for the area considered (NYC). 
Finally, we deem certain features unextractable due to unavailable data at the time of analysis. In this case, we leverage the flexibility of our weight calculation algorithm to eliminate these features from the pool of indicators.

\subsection{Robot Deployments} \label{sec:Deployments}

We devised a case study experiment to qualitatively evaluate our proof-of-concept robotability deployment in New York City. We wished to gain a sense of how well $R$ reflects and helps predict deployment conditions in practice. To this end, we deployed Trashbot \cite{bu_trash_2023}, a differential-drive mobile robot taking the form of a trash barrel, to traverse the parts of the city with very high and very low Robotability Scores. The robot was remotely operated using wizard-of-oz methodology %
by a researcher with a joystick. Trashbot's differential drive movement capabilities include forward motion, backward motion, and rotation in place, akin to a Roomba robot vacuum cleaner. The Trashbot is propelled by recycled hoverboard motors, which provide enough power and speed to navigate the uneven urban terrain in New York City \cite{bu_trash_2023}. We collected video data from the experiment through cameras mounted in the surrounding environment. %

To test if the Robotability Score reflects the suitability of the environment for wheeled robot navigation, we selected four contrasting locations for the experiment: two from the top-scored Census blocks \footnote{Census blocks are a geographic grouping of the U.S. Census Bureau that, in NYC, resemble the area of a single block. They are the smallest grouping for which the Census reports basic demographic data for.} and another two from the lowest-scored blocks. Specifically, we picked one of the highest-ranked and lowest-ranked blocks in the boroughs of Manhattan and Queens. We limited the confounding effect of time-of-day by performing both sets of deployments on two consecutive Thursdays between 3:30 PM and 5 PM. This time aligns with the rush hours in New York City, which is the most challenging time to navigate for urban robots in the city at all locations. 
We selected two wizards to navigate the robot through these locations. The wizards were tasked with navigating a portion of the block at each location and providing feedback on what they found difficult, allowing us to assess the seamlessness of robot navigation at both locations.

\section{Analysis and Results}
\label{sec:results}
In this section, we present analysis and results, including expert insights, computed feature weights, a visualization of $R$ across NYC (as estimated using data from August 2023), and findings from our targeted robot deployments. 

\begin{table*}[h!]
    \begin{tabular}{lllllll}
\toprule
Feature & All & Academia & Industry & Other & NYC POC & Trashbot \\
\midrule
Pedestrian density & \cellcolor{lightgray}0.111 & \cellcolor{lightgray}\textcolor{red}{↓} 0.080 & \cellcolor{lightgray}\textcolor{red}{↓} 0.069 & \cellcolor{lightgray}\textcolor{red}{↓} 0.062 & \cellcolor{lightgray}0.147 & \cellcolor{lightgray}0.173 \\
Crowd dynamics - purpose with which people navigate & \cellcolor{lightgray}0.084 & \cellcolor{lightgray}\textcolor{red}{↓} 0.082 & \textcolor{red}{↓} 0.046 & \textcolor{red}{↓} 0.034 & \cellcolor{lightgray}0.095 & \cellcolor{lightgray}0.119 \\
Pedestrian flow & \cellcolor{lightgray}0.081 & \cellcolor{lightgray}\textcolor{red}{↓} 0.072 & \textcolor{red}{↓} 0.023 & \textcolor{red}{↓} 0.034 & - & - \\
Surface condition & 0.066 & \textcolor{red}{↓} 0.063 & \textcolor{red}{↓} 0.053 & \textcolor{red}{↓} 0.050 & \cellcolor{lightgray}0.092 & \cellcolor{lightgray}0.119 \\
Sidewalk width & 0.062 & \textcolor{red}{↓} 0.058 & \cellcolor{lightgray}\textcolor{ForestGreen}{↑} 0.063 & \cellcolor{lightgray}\textcolor{ForestGreen}{↑} 0.064 & 0.079 & 0.083 \\
Density of street furniture (e.g. garbage, poles) & 0.059 & \textcolor{ForestGreen}{↑} 0.061 & \cellcolor{lightgray}\textcolor{ForestGreen}{↑} 0.060 & \textcolor{red}{↓} 0.058 & 0.076 & 0.090 \\
Intersection safety & 0.057 & \textcolor{red}{↓} 0.051 & \textcolor{red}{↓} 0.024 & \textcolor{red}{↓} 0.019 & 0.066 & - \\
Weather conditions & 0.049 & \textcolor{ForestGreen}{↑} 0.058 & \textcolor{red}{↓} 0.038 & \textcolor{red}{↓} 0.040 & - & - \\
Curb ramp availability & 0.048 & \textcolor{ForestGreen}{↑} 0.055 & \textcolor{red}{↓} 0.044 & \textcolor{ForestGreen}{↑} 0.049 & 0.060 & 0.078 \\
Wireless communication infrastructure (e.g. 5G, IoT, Wi-Fi) & 0.046 & \textcolor{red}{↓} 0.032 & \textcolor{ForestGreen}{↑} 0.054 & \textcolor{ForestGreen}{↑} 0.056 & 0.054 & 0.057 \\
Existence of detailed digital maps of the area & 0.040 & \textcolor{red}{↓} 0.035 & \textcolor{red}{↓} 0.026 & \textcolor{red}{↓} 0.037 & 0.048 & 0.062 \\
Sidewalk / Surface roughness & 0.036 & \textcolor{ForestGreen}{↑} 0.043 & \textcolor{ForestGreen}{↑} 0.052 & \cellcolor{lightgray}\textcolor{ForestGreen}{↑} 0.062 & 0.038 & 0.038 \\
GPS signal strength & 0.034 & \textcolor{red}{↓} 0.031 & \textcolor{ForestGreen}{↑} 0.038 & \textcolor{red}{↓} 0.028 & 0.041 & 0.051 \\
Local attitudes towards robots & 0.032 & \textcolor{ForestGreen}{↑} 0.033 & \textcolor{ForestGreen}{↑} 0.047 & \textcolor{ForestGreen}{↑} 0.046 & - & - \\
Vehicle traffic & 0.031 & \textcolor{ForestGreen}{↑} 0.034 & \textcolor{ForestGreen}{↑} 0.037 & \textcolor{red}{↓} 0.026 & 0.038 & - \\
Traffic management systems & 0.030 & \textcolor{ForestGreen}{↑} 0.032 & \textcolor{ForestGreen}{↑} 0.041 & \textcolor{ForestGreen}{↑} 0.049 & 0.034 & - \\
Slope gradient (i.e. elevation change) & 0.029 & \textcolor{ForestGreen}{↑} 0.038 & \textcolor{ForestGreen}{↑} 0.048 & \textcolor{ForestGreen}{↑} 0.062 & 0.037 & 0.048 \\
Zoning laws and regulation & 0.026 & \textcolor{ForestGreen}{↑} 0.033 & \textcolor{ForestGreen}{↑} 0.038 & \textcolor{ForestGreen}{↑} 0.037 & 0.031 & - \\
Street lighting & 0.019 & \textcolor{ForestGreen}{↑} 0.020 & \textcolor{ForestGreen}{↑} 0.037 & \textcolor{ForestGreen}{↑} 0.023 & - & - \\
Bicycle traffic & 0.017 & \textcolor{ForestGreen}{↑} 0.028 & \textcolor{ForestGreen}{↑} 0.018 & \textcolor{ForestGreen}{↑} 0.021 & 0.020 & 0.025 \\
Proximity to charging stations & 0.015 & \textcolor{ForestGreen}{↑} 0.017 & \textcolor{ForestGreen}{↑} 0.043 & \textcolor{ForestGreen}{↑} 0.043 & 0.018 & 0.023 \\
Bike lane availability & 0.012 & \textcolor{ForestGreen}{↑} 0.016 & \textcolor{ForestGreen}{↑} 0.029 & \textcolor{ForestGreen}{↑} 0.037 & 0.013 & 0.016 \\
Surveillance coverage (CCTV) & 0.012 & \textcolor{ForestGreen}{↑} 0.014 & \textcolor{ForestGreen}{↑} 0.035 & \textcolor{ForestGreen}{↑} 0.034 & 0.014 & 0.019 \\
Existence of shade (e.g., trees) & 0.008 & \textcolor{ForestGreen}{↑} 0.014 & \textcolor{ForestGreen}{↑} 0.037 & \textcolor{ForestGreen}{↑} 0.030 & - & - \\
\bottomrule
\end{tabular}

    \caption{Robotability feature weights in aggregate, from different expert demographics, and for the minified configuration of the score used for the NYC TrashBot \cite{bu_trash_2023} deployment. The ``Other'' column includes Accessibility and Urban Planning experts. Cells highlighted in grey represent the top 3 features as ranked per each expert group. Up and down arrows indicate how the weight compares to the aggregate weight set. The ``NYC POC'' column depicts the weights used to generate \autoref{fig:rs-nyc-map}.}
    \label{tab:weights}
\end{table*} 

\subsection{Expert Interviews}

We interviewed a total of 10 experts. Four work in robotic navigation in Academia, two are involved in robotics in Industry, one in the field of Urban Planning, one in the field of robotics legislation, one in the field of accessibility in Academia, and one in the field of accessibility in industry. 

The experts' professional backgrounds impact the feedback received. For example, we identified different value sets concerning robot navigation - experts using ethnographic methods in their research are more likely to prefer ``interactional'' or ``crowd-dependent'' features (e.g., local attitudes towards robots). In contrast, robot technicians tended to focus more on technical aspects concerning navigation (e.g., GPS signal) or the built environment. 

Experts shared personal experiences with robot navigation, namely the challenges of environmental changes (street furniture), sidewalk conditions making robot navigation infeasible, and instances of robot bullying. Experts also pointed out the analogous nature of robotability for wheeled robots and Accessibility. Some experts, namely those with an urban planning background, referred to the importance of clarification of the \textit{intended purpose} of $R$ -- it should not be a measure that intends to skew urban planners towards building a \textit{city for robots}; instead, the Robotability Score is a surveying metric which can inform roboticists about the feasibility of deploying a robot in a specific urban location.
 
The experts piloted the survey and offered suggestions for improving its flow and clarity. We decided to add a ``feature cheat sheet'' to our survey, for participants to consult while taking the survey. We finalized the feature list after the interview stage of our study, shown below (Table \ref{tab:weights}).%

\subsection{Survey and Robotability Feature Weights}

We obtained a total of 47 survey answers (68.0\% from a background of robotics in Academia, 17.0\% from robotics in Industry, 6.4\% respondents from an Urban Planning background, 4.3\% Accessibility experts, and 4.3\% others), with an average professional experience of 8.3 years. We collected a total of 1587 pairwise comparison answers. To evaluate the consistency of responses, we calculate answer \textit{transitivity} (e.g., if A is preferred over B and B over C, then A should be preferred over C). For \textit{intra-rater transitivity}, there are no transitivity violations for individual sets of answers. Across the dataset (\textit{inter-rater transitivity}) we find 616 transitivity violations, approximately $30\%$ of the 3-way combinations of items.

\subsubsection{Weights}
The full list of features considered relevant to calculate the Robotability Score comprises 24 indicators. %
In Table \ref{tab:weights}, we present the weight sets for robotability, as reported in aggregate, from academic experts, industry experts, and other survey-takers. This representation allows us to analyze differences in values for different demographics of experts, in line with identifying matrix transitivity.

\begin{landscape}
\begin{table}
    \begin{tabular}{p{5cm}lp{7cm}p{5.5cm}}
\toprule
\bf Feature Name & \bf Pol. & \bf Example Schema & \bf Dataset / Proxy \\
\midrule
Pedestrian density & $-$ & 0 = high density, 1 = low density &  Derived from $D$, see \ref{sec:dashcam} \\
Surface condition & $+$ & 0 = potholes, 1 = major cracks, 2 = minor cracks, 3 = smooth &  NYC OpenData - Sidewalk Scorecard Ratings \cite{nyc_opendata_scorecard_2024} \\
Crowd dynamics  & $-$ & 0 = emergency, 1 = commuting, 2 = shopping, 3 = leisure & Derived from NYC Zoning Map (ZOLA) \cite{nyc_department_of_city_planning_zola_2024} \\
Intersection safety & $+$ & 0 = no infrastructure, 1 = basic crosswalks, 2 = zebra crossings with signals, 3 = advanced intersection management & Various safety-oriented infrastructural components, described in \autoref{sec:feature-computation} \ \\
Sidewalk width & $+$ & 0 = narrow, 1 = wide & Inferred from NYC Plainometric Database \cite{nyc_opendata_nyc_2024} \\
Density of street furniture & $-$ & 0 = high density, 1 = moderate density, 2 = low density, 3 = no furniture & Computed via aggregation of several public street furniture datasets, see \autoref{sec:feature-computation} \\
Wireless communication infrastructure & $+$ & 0 = no infrastructure, 1 = basic Wi-Fi, 2 = 5G, IoT & FCC Mobile LTE Coverage Map \cite{us_federal_communications_commission_mobile_2024} \\
Curb ramp availability & $+$ & 0 = no curb ramps, 1 = curb ramps at some intersections, 2 = curb ramps at most intersections, 3 = curb ramps with tactile paving & NYC OpenData - Pedestrian Ramp Locations \cite{nyc_opendata_pedestrian_2024} \\
Slope gradient & $-$ & 0 = high slope, 1 = no slope & Derived from NYC OpenData - 1-foot Digital Elevation Map \cite{nyc_opendata_1_2024} \\
GPS signal strength & $+$ & 0 = no signal, 1 = very strong signal & Assume 1 citywide \\
Sidewalk / Surface roughness & $-$ & 0 = gravel, 1 = cobblestones, 2 = smooth asphalt &  Assume 1 citywide \\
Existence of detailed digital maps of the area & $+$ & 0 = no digital maps, 1 = highly detailed maps & Assume 1 citywide \\
Zoning laws and regulation & $+$ & 0 = no regulations, 1 = moderate stringency, 2 = high stringency & NYC OpenData - VZW Speed Limits \cite{nyc_opendata_vzv_speed_2024} \\
Traffic management systems & $+$ & 0 = no system, 1 = basic signage, 2 = street lights, 3 = advanced traffic control & Derived from various built features of traffic intersections, see \autoref{sec:feature-computation}  \\
Vehicle traffic & $-$ & 0 = very high traffic, 1 = very low traffic & Derived from $D$, see \ref{sec:dashcam} \\
Bicycle traffic & $-$ & 0 = very high traffic, 1 = very low traffic & Derived from $D$, see \ref{sec:dashcam}\\
Proximity to charging stations & $+$ & 0 = no charging stations nearby, 1 = multiple charging stations in proximity & CitiBike Station Locations \cite{lyft_citibike_2024} \\
Surveillance coverage (CCTV) & $+$ & 0 = no coverage, 1 = low coverage, 2 = moderate coverage, 3 = high coverage & Decode Surveillance NYC Dataset \cite{noauthor_inside_2022} \\
Bike lane availability & $+$ & 0 = no bike lanes, 1 = bike lanes on most roads, 2 = separated bike lanes on all roads & NYC OpenData - Bike Routes \cite{nyc_opendata_new_2024} \\
Local attitudes towards robots & $+$ & 0 = highly negative, 1 = somewhat negative, 2 = neutral, 3 = somewhat positive, 4 = highly positive & \cellcolor{gray!25} Data not available \\
Weather conditions & $+$ & 0 = ice, 1 = snow, 2 = puddles, 3 = leaves, 4 = clear & Determined in real-time \\
Street lighting & $+$ & 0 = no lighting, 1 = excellent lighting & \cellcolor{gray!25} Data not available \\
Existence of shade & $+$ & 0 = no shade, 1 = extensive shade & \cellcolor{gray!25} Data not available \\
Pedestrian flow & $-$ & 0 = very high flow, 1 = very low flow & \cellcolor{gray!25} Data not available \\
\bottomrule
\end{tabular}

    \caption{The compiled feature list for $R$. We additionally include the assigned polarity, an example categorization scheme, and the dataset or proxy used to compute the feature in our proof-of-concept deployment.}
    \label{tab:indicators}

\end{table}
\end{landscape}

\subsection{Feature Extraction}

In \autoref{tab:indicators}, we present the finalized list of features that compose $R$, along with polarity, a point-based system provided to survey takers as an example of the feature use, and data proxies used to compute the features in New York City. "$+$" indicates positive polarity, and "$-$" indicates negative polarity. The point-based examples provided in parenthesis illustrate the polarity takes effect in calculating the impact of each feature.

For New York City (NYC), we consider 19 out of 24 possible indicators. Street lighting, shade availability, pedestrian flow, and human attitudes towards robots are excluded due to unavailable data. Also excluded is weather conditions, a feature meant to be considered in real-time. We assume GPS strength, surface roughness, and the existence of detailed digital maps of the area to be 1 citywide; our rationale behind this analysis decision can be found in Appendix \ref{sec:feature-computation}.

The specific implementation of our feature computation pipeline can be found in the study repository.

\subsection{Robotability Score in NYC}
\begin{figure}[h!]
    \includegraphics[width=0.48\textwidth]{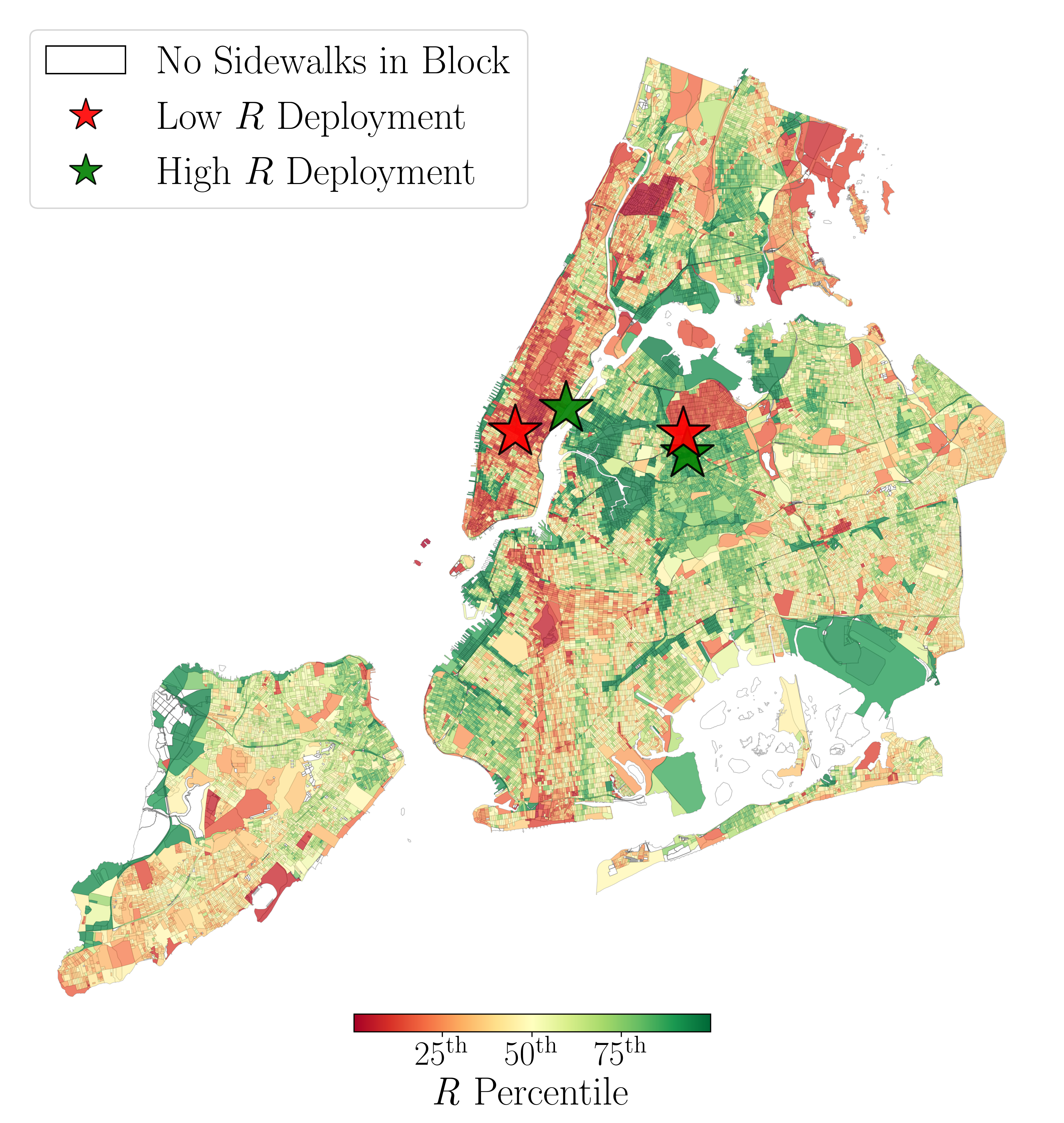}
    \caption{Robotability Score distribution in New York City. Blocks that are colored white indicate a lack of sidewalks, or a lack of dashcam data to estimate features. Red dots indicate the locations of our two deployments in areas of lowest $R$, and green dots indicate the locations of our two deployments in areas of highest $R$. The weights used to generate this distribution are in the ``NYC POC'' column of \autoref{tab:weights}. A larger map, displaying the robotability distribution from the perspective of TrashBot's requirements, can be viewed in \autoref{fig:rs-nyc-trashbot}.}%
    \Description{Map of different boroughs of New York City, colored by the Robotability Score distribution. The plot utilizes a red-to-green gradient coloring scheme; red areas are low robotability, and green areas are high robotability. Blocks that are colored white indicate a lack of sidewalks, or a lack of dashcam data to estimate features. These are mostly present in Staten Island. Red dots indicate the locations of our two deployments in areas of lowest Robotability Score, and green dots indicate the locations of our two deployments in areas of highest R, which are described in the main text.}
    \label{fig:rs-nyc-map}
\end{figure}

In \autoref{fig:rs-nyc-map}, we show the citywide distribution of $R$, estimated using data from August 2023 and a weight set recalculated to the data-available feature set, which can be found in \autoref{tab:weights}. For datasets that provide start dates or installation dates, we remove data that originate after the cut-off period of August 31st, 2023. \autoref{fig:rs-nyc-map} is colored by census block.

\setlength{\tabcolsep}{0.1em} %
\begin{figure*}[h!]
    \renewcommand*{\arraystretch}{0.5}
    \centering
    \begin{subfigure}[t]{\textwidth}
        \centering
        \begin{tabular}{ll}
            \includegraphics[width=0.5\textwidth, height=0.15\paperheight]{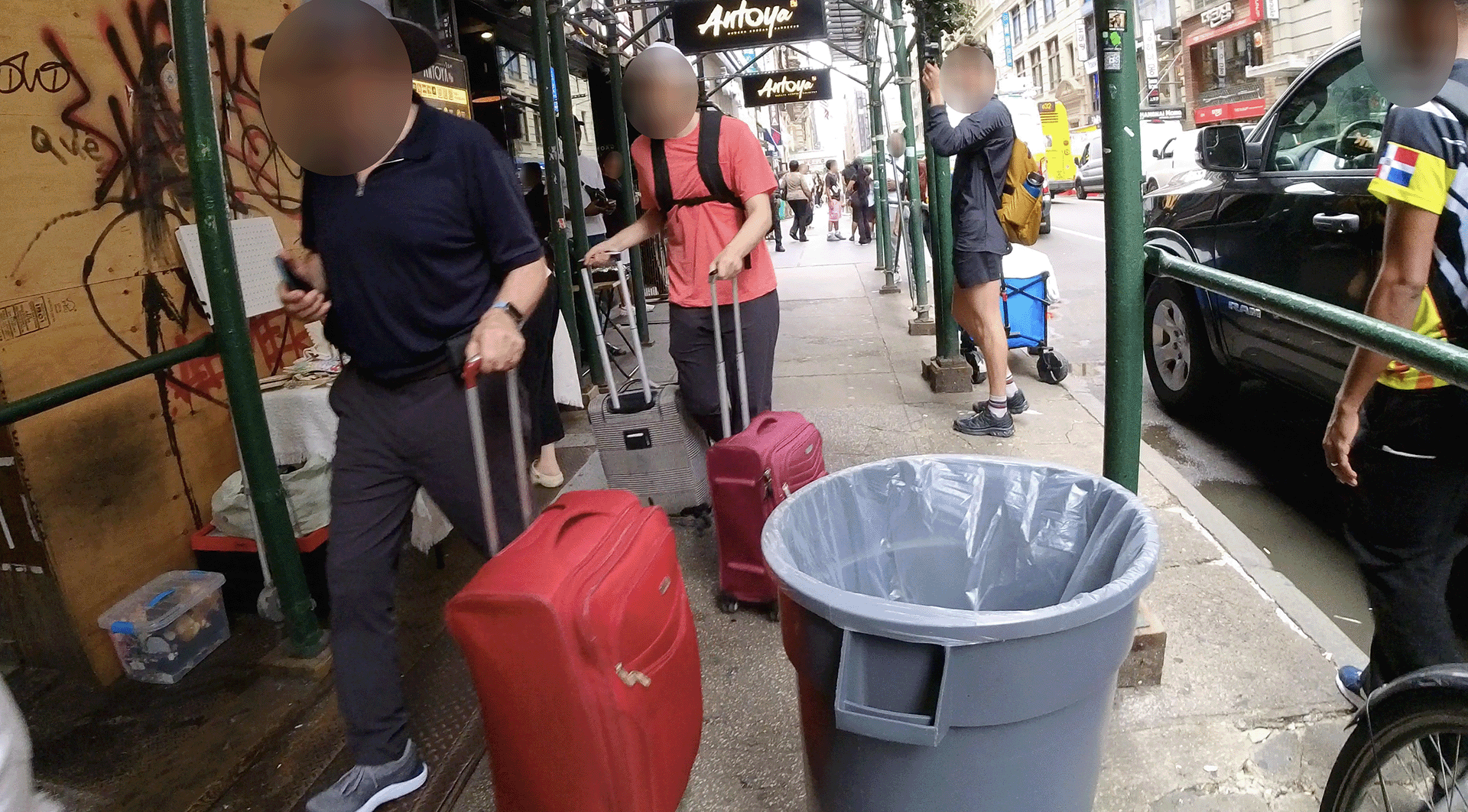} & 
            \includegraphics[width=0.5\textwidth, height=0.15\paperheight]{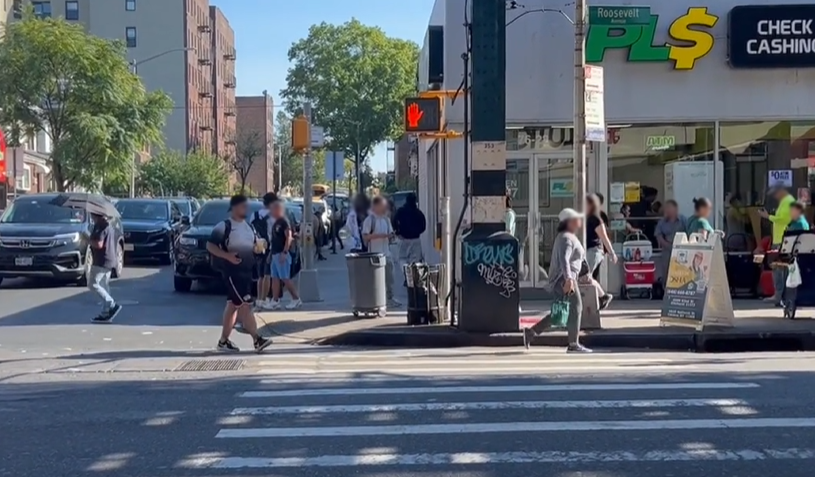} \\
            \includegraphics[width=0.5\textwidth, height=0.15\paperheight]{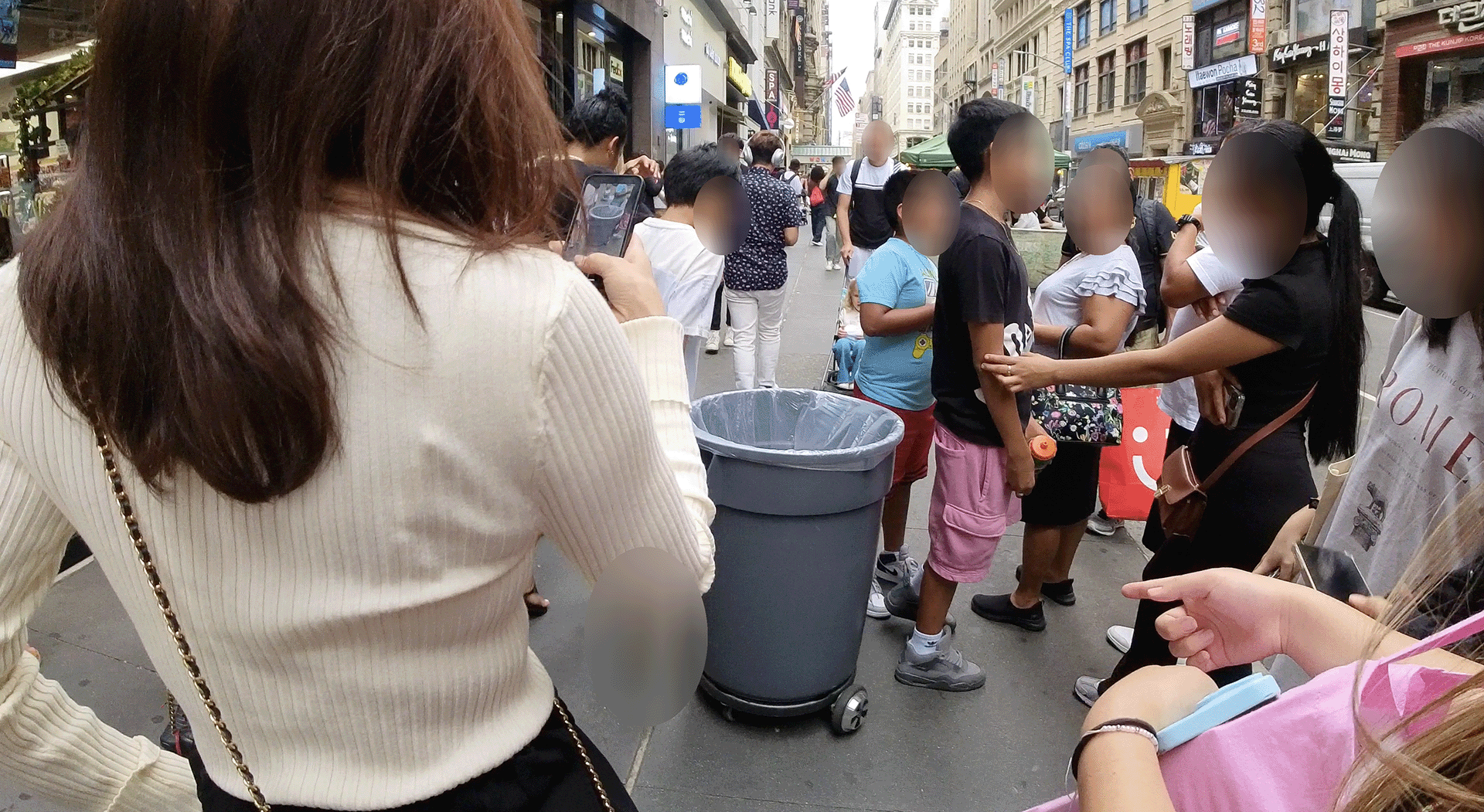} & 
            \includegraphics[width=0.5\textwidth, height=0.15\paperheight]{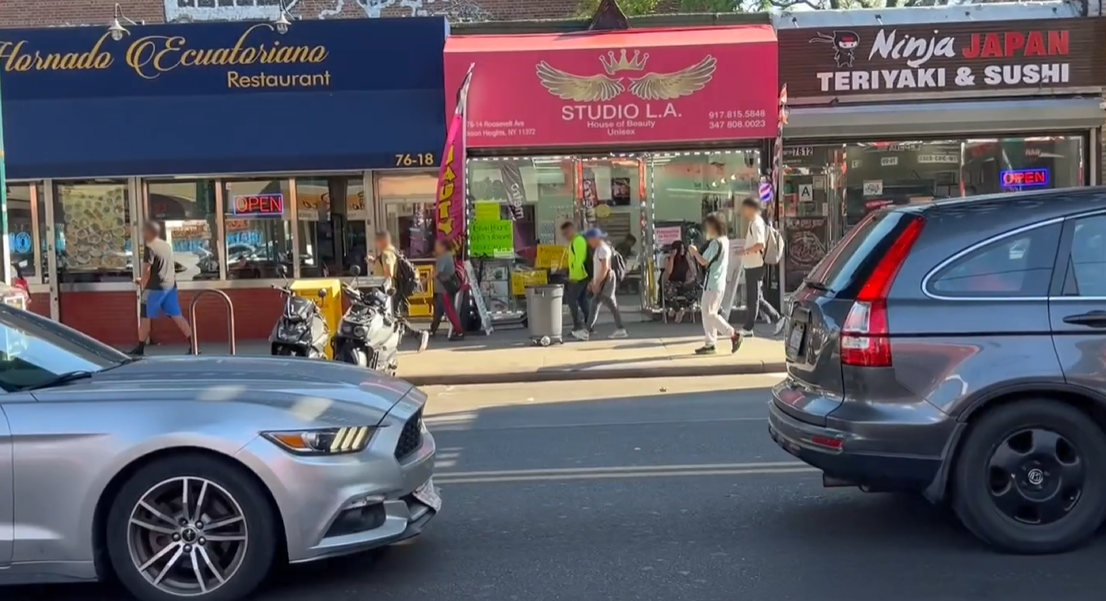} \\
        \end{tabular}
        \caption{Deployment in Area of Lowest $R$}
        \label{fig:lowest-robotability}

    \end{subfigure}
    \space 
    \begin{subfigure}[t]{\textwidth}
        \centering
        \begin{tabular}{l}
            \includegraphics[width=0.5\textwidth, height=0.15\paperheight, trim={0 0 0 2cm}, clip]{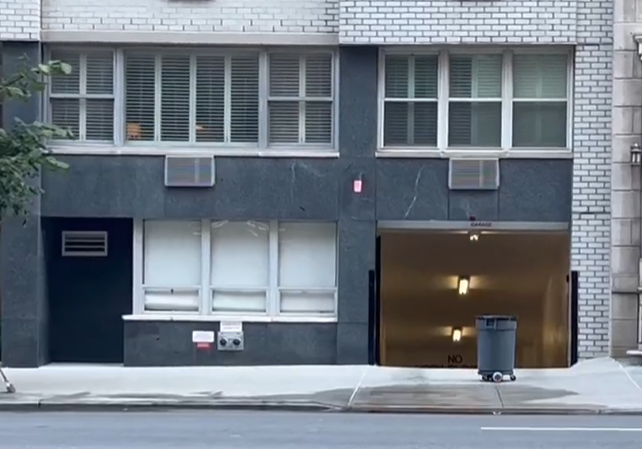}
            \includegraphics[width=0.5\textwidth, height=0.15\paperheight, trim={0 0 2cm 2cm}, clip]{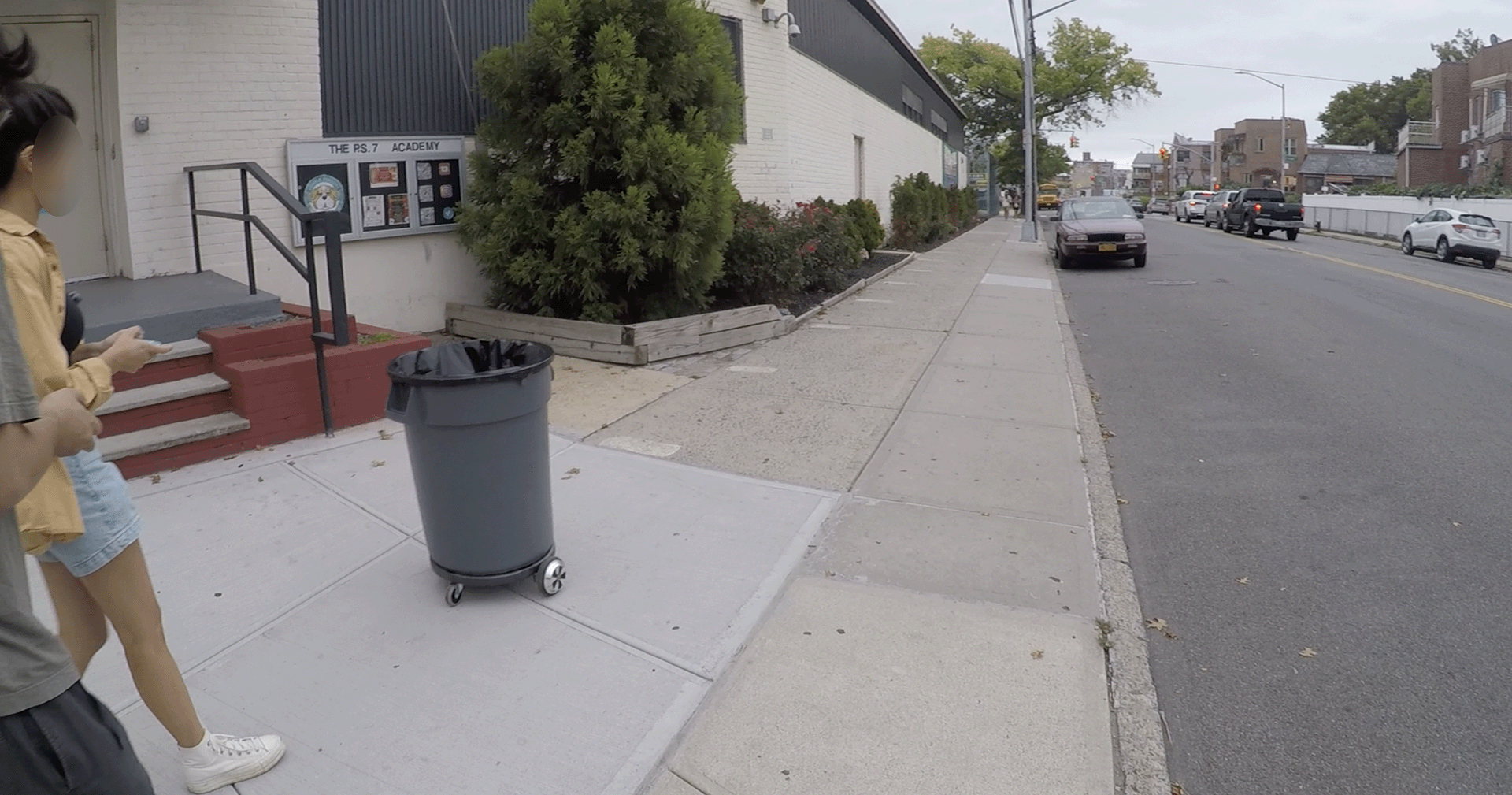} \\
        \end{tabular}
        \caption{Deployment in Area of Highest $R$}
        \label{fig:highest-robotability}

    \end{subfigure}
\caption{Trashbot deployments in two of the highest and two of the lowest robotability neighborhoods in NYC.}
\Description{A total of 6 pictures are displayed. Four pictures (Figure a) display Trashbot, a trashcan-on-wheels robot, in high pedestrian density areas, surrounded by people of different ages, corresponding to areas of high Robotability Score. The two bottom pictures (Figure b) display empty, large sidewalks which are smooth, illustrating Trashbot navigation in two neighborhoods of low robotability.}
\label{fig:deployments}
\end{figure*}
\setlength{\tabcolsep}{0.5em} %

\subsection{Robot Deployment}

In our case study deployment, we deliberately excluded the following factors in the computation of the Robotability Score: traffic management, zoning laws, shade, intersection safety, vehicle traffic, and bike lane availability. This decision was based on the specific context of our robot, TrashBot -- a trash barrel robot designed to navigate sidewalks and plazas to collect trash, not intended to cross intersections or navigate on streets with car traffic. We recalculated the weight set based on available feature data and robot specifications, as can be seen on \autoref{tab:weights}. The TrashBot-specific distribution of $R$ is visualized spatially in \autoref{fig:rs-nyc-trashbot}. Below, we describe the deployment sessions briefly, which are shown pictorially in \autoref{fig:deployments}. We selected four locations across the two boroughs of Manhattan and Queens: two in the top 10\% of $R$, and two in the bottom 10\% of $R$; locations are arbitrarily influenced by physical ease of deployment (ie. while Times Square would be interesting, we would have run into logistic problems during the deployment).

\subsubsection{High Robotability Score}

\paragraph{Elmhurst, Queens}
Elmhurst presents an urban layout where stores are dispersed throughout the neighborhood, predominantly filled with residential buildings. The neighborhood features wide sidewalks with little to no foot traffic.

Our robot deployment in Elmhurst proceeded smoothly, even during New York City's rush hour. We encountered only two groups of people when navigating around an entire block. The primary challenge we faced was the under-maintained condition of the sidewalks, which was unexpectedly exacerbated by ongoing construction nearby.

\paragraph{Sutton Place, Manhattan}
Sutton Place is a quiet neighborhood in Manhattan near the East River. It is mostly a residential area with little commercial stores. Since it is far from major subway stations, it has little foot traffic, and most passersby are local residents. The peaceful environment makes it one of the most suitable neighborhoods in busy Manhattan for robot deployments. 

During our deployment, the wizards rarely ran into any pedestrians, even if it was around rush hour in New York City. The pavement was clean and flat, with a water pipe being the only obstacle throughout the deployment. 

\subsubsection{Low Robotability Score}
\paragraph{Herald Square, Manhattan}
Herald Square in Manhattan is located in the vicinity of Koreatown. The area is renowned for its array of restaurants, bakeries, and grocery stores that line the streets, drawing substantial foot traffic throughout the day. While the sidewalks are well-maintained, our deployment faces significant challenges due to the dense foot traffic, with people moving in both directions on both sides of the sidewalks. Additionally, the presence of numerous street vendors scattered along the sidewalks further complicates robot navigation. 

When we deployed our robot here, the robot was quickly surrounded by bemused pedestrians. The wizards lost their line of sight on the robot multiple times due to the dense foot traffic. The robot had a hard time finding a path forward. The entropy of pedestrian flow was heightened by street vendors setting up tables on both sides of the street, aggravating the difficulty in maneuvering the robot.  

\paragraph{Jackson Heights, Queens}
Jackson Heights stands out as one of New York City's most varied neighborhoods, drawing in lots of residents and visitors from many cultures. This neighborhood also acts as a crucial transportation hub, linking the west and east sides of borough of Queens.

Robot navigation through Jackson Heights faces challenges due to the heavy pedestrian traffic during the day, a nearby public transportation hub, relatively narrow sidewalks, and intricate road infrastructure. Additionally, stores frequently extend their displays onto the sidewalks, and food vendors often set up their stands along the curbside, further congesting the already limited space. During our deployment, the wizards struggled to constantly navigate the robot along with dense crowds and avoid temporary stands, signaling Jackson Heights as one of the most challenging neighborhoods for robot deployments.

The alignment between the conditions met at deployment and the Robotability Score grounds our approach, demonstrating its capacity to be tailored to different robotic applications and environments. 
At the same time, we notice the sharp need for more accurate and expansive data streams. 

\clearpage

\section{Discussion}
\label{sec:discussion}

This work introduces the Robotability Score, a novel and future-facing measure of how the urban environment supports robot navigation. Through a multi-stage process and by incorporating expert knowledge, we compile a list of features which compose robotability and weigh them in their impact on wheeled robot navigation. The features concern both existing infrastructure and dynamic human considerations, such as local attitudes towards robots. There are two observations to be made on the feature selection and weighting methodology. First, we asked experts and survey literature to find which features should be included independently of whether the features are already available. As public mobile robots become more frequent occupants of urban streets \cite{tiddi2019rci}, the demand (and hence supply) for these data will increase. $R$ can meld to existing data availability without compromising its usefulness, as comparisons across areas may still be drawn, as can be seen in the use case of New York City (\autoref{fig:rs-nyc-map}). Second, the weights extracted using the pairwise comparison methodology might be skewed towards the background and professional experience of the respondents, e.g., robot technicians might highlight infrastructure features since that concerns the design of navigation hardware on the robot, whereas ethnographic experts highlight interactive features (how the robot interacts with humans) as playing an important role. This is shown in our results through transitivity violations and also when recalculating weights conditionally to the experts' professional background (\autoref{tab:weights}). It would have been informative to estimate the weight distribution from the set of urban planning ($N=3$) and accessibility ($N=2$) experts, but we were unable to secure a large enough number of subsamples during the study period. 
A more generalizable score would require a higher and more balanced number of expert responses to the survey.

In addition to securing a larger sample of expert responses for weight configuration, the representativeness of robotability to real environments can be improved with more, and better, data. Traffic densities, including pedestrian, vehicle, and bicycle volumes, are derived from large-scale dashcam data that is not widely available globally. Future work might explore the use of cell phone mobility data as a more available proxy for traffic-involved features of $R$. \cite{caceres_traffic_2012,wang_estimating_2021}

Several features are incomputable due to unavailable data. Shade availability could be computed via a tool called ShadeMap \cite{shade_map_about_2024}, but we are unable to extract an area for batch processing as large as NYC. For the scope of this work, we deem pedestrian flow data incomputable at the city-wide level. Some future ideas for computation of this feature include using the density of points of interest (POIs) as a proxy, measuring direct pedestrian flow through an offering like the smart cameras made by Numina \cite{numina}, or doing time-based averaging on our pedestrian traffic count data. Weather conditions are the most temporally sensitive feature included in robotability, and thus are to be computed in real-time or at the time of deployment. In our deployment, we assume a citywide value of 1, as there are no rainstorms in the area during any deployment period. Separately, the street lighting indicator is also deemed incomputable due to a lack of data in NYC; while it is possible to train a supervised vision model to detect street lights, we deem the overhead of this out of scope, especially relative to the overhead required to compute other indicators of $R$. Lastly, the \textit{local attitudes towards robots} indicator is particularly forward-facing; even if we were to survey residents of NYC about their sentiments of co-navigating with robots, they likely would have spent little time internalizing this concept; simply put, robots are not presently an everyday encounter for the average pedestrian.

An important characteristic of our deployment and computation of $R$ is that no online sensing is involved in the robot's navigation; that is, information from onboard sensors on TrashBot was not utilized to inform, influence, or diagnose navigation. This is an important avenue of future work that would pair well with commercial adoption of robotability.

For our proof-of-concept deployment in New York City, we focus on qualitative evaluation metrics, with TrashBot as a prototype robot. While the Robotability Score is undoubtedly a metric downstream of our present society, environments do exist where an ecologically-valid deployment is possible. In the United States, various college campuses are piloting food delivery robots \cite{garwood_delivery_2022}. In these cases, a sample objective for the robot would be to deliver the goods as quickly, safely, and politely as possible. One can imagine a deployment based on a formal experiment, where one group of robots utilizes a legacy navigation routine and the other makes use of weighted navigation from the Robotability Score. One other drawback of our deployment is the unknown distribution shift between August 2023, when we have dashcam data for and thus when we compute $R$ over, and August 2024, when we pursue our proof-of-concept deployments. 

A unique opportunity in the further development of robotability that is infeasible when crafting a metric for humans comes from the fact that robotic capabilities can be directly quantified. In our tailored proof-of-concept for TrashBot, we know exactly the slope gradient that the onboard motor could handle, the wireless connectivity requirements, and the width of the sidewalk required for navigation. Of course, ambiguity still exists for indicators that involve interaction with humans or human-powered vehicles. Even so, this score is highly adaptable -- different sets of weights can emerge based on different classes of robots. Not only may the weights change, but the polarity as well; for example, a solar-powered delivery robot might actually be disincentivized to seek shade. This said, one limitation arises from the case of needing to consider a \textit{new} feature in a configuration of robotability; in this case, new pairwise comparisons would need to be gathered with the set of new features $F_n+l$ against the set of existing features $F_n$.

Robotability is not only extendable to different robot architectures, but also different types of human development. We focus on cities in this work due to their complexity and density, but subsets of our indicator list can apply to suburban, rural, and otherwise different civilization patterns.
We acknowledge that New York City represents a rare example of urban environments with extensive data availability; not all features are currently measured in other urban areas. As urban data sources continue to grow, more environments will become viable for robotability assessment. Future research will explore leveraging machine learning techniques to extrapolate missing features in data-scarce areas, potentially enabling broader urban robotability mapping across different cities, analogously to recent developments in accessibility classification  \cite{LiuTowardsASSETS2024, koh2023routing}. A dynamic and automatically updated version of this score could also contribute to more efficient and explainable robot navigation routing \cite{suresh2023robotnavigationriskycrowded}. For example, it can allow roboticists to account for sudden changes in weather conditions or pedestrian density.

Finally, it is crucial to note that the Robotability Score is not intended to replace or compete with established metrics such as walkability or accessibility. Rather, it serves as a complementary tool that provides additional insights into the urban landscape. Features that improve walkability and accessibility in public spaces (e.g., PROWAG \cite{accessboard_prowag}) have been long fought for and are still actively being pursued. Sharing urban streets with robots should not affect the accessibility of these spaces and engineering robot navigation must take these considerations into account \cite{bennet2021crowded}, in efforts to achieve harmonious co-existence. There is a growing number of tools that assess the accessibility of urban spaces, many of which emerge from community-driven efforts. Examples of this are Roll Mobility \cite{rollmobility}, a community-driven tool that reviews the accessibility of restaurants, public spaces, among others, in efforts to promote inclusion. %
Many educational institutions have also put forth tools, such as AccessMap \cite{bolten2019access} (an open-source tool that allows for route planning according to the user's specific mobility needs), or Project Sidewalk \cite{saha2019sidewalk}, which relies on efficient crowdsourcing for data collection on accessibility. In line with these projects, the Robotability Score is made open-source so it can be further developed for wider use.

In this work, we develop a novel, anticipatory, and extensible metric for improved robot navigation. Our framework underlying the development of robotability demonstrates how to operationalize expert knowledge and synthesize it with wide-ranging datasets to finely model a real-world urban phenomenon. We are hopeful that robotability will guide the design of both robots and inform interactions that preserve harmony upon the introduction of robots into city streets.

\begin{acks}
We thank our experts interviewed for their valuable insights into how to think about robotability. We thank Nexar, Inc. for research support and collaboration, and notably for providing the dashcam data used to generate traffic estimates in our proof-of-concept deployment. We thank Nik Martelaro, Gabriel Agostini, Ilan Mandel, and Hauke Sandhaus for project feedback. We thank Stacey Li, Melina Tsai, and Albert Han for serving as expert wizards in our NYC deployments. We thank Meli Harvey for publishing materials from her \textit{Sidewalk Widths NYC} project, which we used to bootstrap our generation of the sidewalk network. This work was funded in part by an NSF FRR grant, NSF IIS 2423127. Lastly, we thank the Urban Tech Hub and the Digital Life Initiative at Cornell Tech for longstanding research support. 
\end{acks}

\balance
\bibliographystyle{ACM-Reference-Format}
\bibliography{sources/matt-references, sources/bib}

\clearpage 

\appendix 

\section{Robotability Feature Computation}
\label{sec:feature-computation}

In this Appendix section, we details the dataset(s) and/or proxies used for each feature in our proof-of-concept computation of robotability in New York City. Unless otherwise noted, the coverage of each dataset is up-to-date relative to our data cutoff period of August 31st, 2023.  

\subsection{Pedestrian density}
We compute pedestrian density via the methods described in \ref{sec:dashcam}.

\subsection{Surface condition}
We measure the surface condition indicator through the \textit{Sidewalk Scorecards} system that NYC has designed and implemented across its 50+ Community Districts \cite{nyc_opendata_scorecard_2024}. The metric we use here is the percent of sidewalks deemed acceptable by an inspector, in a Community District. Community Districts are much larger than census blocks, which we use in our analysis; this is a limitation of this data source. There are other limitations rooted in the subjectivity of human ratings, etc. 

\subsection{Crowd dynamics}
In our proof-of-concept deployment, we model crowd dynamics by considering the zoning of the immediately surrounding area. Commercial zones are given the lowest score, followed by residential zones, followed by manufacturing zones. 

\subsection{Intersection safety}
We represent intersection safety in our proof-of-concept deployment via motor vehicle collision data published on NYC OpenData \cite{nyc_opendata_motor_2024}. We filter this data for collisions that involved vehicles and pedestrians. Our logic for using vehicle-pedestrian collisions as a proxy for $R$'s intersection safety indicator is that if an intersection is unsafe for pedestrians, it is also likely unsafe for wheeled robots. 

\subsection{Sidewalk width}
NYC government provides a publicly-available planimetric database \cite{nyc_opendata_planimetric_2024} that contains computationally-parsable sidewalk centerlines, and an estimate of width. However, the sidewalks in this database tend to be organized at the block level, and so aren't accurate in a situation where two perpendicular sidewalks have different widths (ie. a major avenue perpendicular to a side street). We follow the methodology of \cite{moynihan_mapping_2020} to parse the sidewalk database into individual sidewalk segments, each with their own inferred width. 

Vetted and published investigative data projects have used this data to determine sidewalk widths \cite{fadulu_think_2024} \cite{moynihan_mapping_2020}. For our purposes, where integer precision would be sufficient, this dataset is acceptable. 

\subsection{Density of street furniture}
We compute street furniture densities via a two-prong approach. We first take walks around neighborhoods in Manhattan, Queens, and Brooklyn, and generate a representative list of potential, built sidewalk impediments. The street furnitures considered and processed to compute this are provided in \autoref{tab:street-furniture-features}. We take every type of street furniture and overlay them on the segmentized sidewalk graph, with a sidewalk point being matched to a street furniture if they are within approximately 7.5 meters (about 25 feet).

\subsection{ Wireless communication infrastructure}
We use the Federal Communications Commission's national broadband map to represent the wireless communication infrastructure feature of $R$. This resource is available nationwide and maps 4G LTE and 5G download and upload speeds at fine granularity \cite{us_federal_communications_commission_mobile_2024}. For TrashBot, which connects to the internet via a cellular network-powered router, if both the 4G download and upload speed at a node in the graph exceed 10 MB/s, this feature takes a value of 1; otherwise, the requirement fails and the feature takes a value of 0.

\subsection{Curb ramp availability}
The curb ramp availability indicator is computed via a spatial merge between the nodes of our sidewalk graph and the NYC OpenData dataset, "Pedestrian Ramp Locations" \cite{nyc_opendata_pedestrian_2024}.

\subsection{Slope gradient}

We compute the slope gradient using an in-house computation described in \ref{sec:slope-gradient-calculation}. The slope gradient is directly computable using elevation data provided by  the 1-Foot Digital Elevation Model of NYC \cite{nyc_opendata_1_2022}. 

\subsubsection{Computing the slope gradient}
\label{sec:slope-gradient-calculation}
In order to estimate the uphill or downhill decline between each pair of segmentized sidewalk points, we need to develop an optimized algorithm. There are \numSidewalkPoints unique points on our segmentized sidewalk graph, and computing the slope gradient for each point requires considering adjacent neighbors. We model the computation of the slope gradient for each point as follows.

Let \( p \) be a point on the segmentized sidewalk graph, and \( P \) be the set of all points on the graph. The elevation of a point \( p \) is denoted by \( p_e \), computed via the sampling of a topology raster file. The slope gradient at a point \( p \) is represented by \( \nabla_p \). \( K \) is the maximum number of neighbors to consider, and \( D \) is the maximum distance within which a point is considered a neighbor.

The set of neighbors \( N_p \) for point \( p \) is defined as follows:

\[
N_p = \{ p' \in P \mid d(p, p') \leq D \}
\]

where \( d(p, p') \) is the distance between points \( p \) and \( p' \).

To limit the number of neighbors to the closest \( K \), we define \( N_p^{(K)} \) as the set of the closest \( K \) points to \( p \) from \( N_p \):

\[
N_p^{(K)} = \text{argmin}_{p' \in N_p} \left\{ d(p, p') \right\}_{K}
\]

The slope gradient \( \nabla_p \) at point \( p \) can then be computed using the elevations of the points in \( N_p^{(K)} \):

\[
\nabla_p = \frac{1}{|N_p^{(K)}|} \sum_{p' \in N_p^{(K)}} \left| \frac{p'_e - p_e}{d(p, p')} \right|
\]

This equation takes the absolute value of each individual slope gradient before averaging, ensuring that only the closest \( K \) neighbors are considered, constrained by the maximum distance \( D \). We take absolute value as we consider a slope too \textit{uphill} or too \text{downhill} as equally detrimental to robotability. This is an assumption.

\subsection{GPS signal strength}
GPS signal strength is modeled at 1 citywide in our proof-of-concept deployment. This is an assumption that may not hold in dense urban environments, where GPS can be blocked by tall buildings \cite{liu_visual_2021}. Nonetheless, as we wizard the robot in our deployments, GPS is superfluous.

\subsection{ Sidewalk / Surface roughness}
We assign a uniform value of 1 to the sidewalk roughness indicator, as we assume a citywide uniform sidewalk material of concrete. We make this assumption due to a lack of data, which is not currently published on NYC OpenData. Computing this feature in-house is possible with a separate supervised computer vision model, but the overhead of this is out-of-scope for this analysis. 

\subsection{Existence of detailed digital maps of the area}
Due to NYC's extensive digital infrastructure and data availability, we assume this feature to be 1 citywide. In a rural area or less-digitized urban center, further processing design of this feature would be necessary. 

\subsection{Zoning laws and regulation}
We model zoning laws and regulations via considering the vehicle speed limits across New York City. The speed limit across the city is largely 25 miles per hour. We source our data from NYC OpenData, where speed limits are published as a part of the Vision Zero traffic safety initiative \cite{nyc_opendata_vzv_speed_2024}.

\subsection{Traffic management systems}
We compute the traffic management feature in our proof-of-concept implementation of $R$ by aggregating various traffic safety or traffic calming infrastructure features reported in NYC's Vision Zero traffic safety initiative \cite{nyc_opendata_vzv_speed_2024}. If a piece of traffic safety infrastructure is nearby a node in the sidewalk graph, the value of this feature increases by 1. The process iterates for each feature presented in \autoref{tab:traffic-management-features}.

\subsection{Vehicle traffic}
We compute the vehicle traffic feature via the methods described in \ref{sec:dashcam}.

\subsection{Bicycle traffic}
We compute the bicycle traffic feature via the methods described in \ref{sec:dashcam}.

\subsection{Proximity to charging stations}
While robot charging stations do not presently exist in New York City, we do identify another system of networked docks: CitiBike stations \cite{lyft_citibike_2024}. We choose to use the distribution of CitiBike stations as hypothetical ``charging'' stations for robots, as they frequently see bikes move between stations and are dispersed with sizable consideration for the existing urban morphology \cite{broderick_identifying_2015}. Then, for each node in our sidewalk graph, we find the distance to the nearest ``charging'' station. 

\subsection{Surveillance coverage (CCTV)}
The surveillance coverage indicator is directly proxied by a dataset from the Decode Surveillance NYC lab \cite{noauthor_numina_nodate}. This dataset includes over 25,000 surveillance camera instances collected from crowdsourced annotations during 2022. A limitation of this data set is that it may not include all known cameras, and more cameras may have been installed since 2022.

\subsection{Bike lane availability}
We estimate bike lane availability via the NYC OpenData Bike Routes dataset, which provides a parsable map of different bike transportation ``facilities'' throughout NYC. The facilities come in three classes, ranging from shared with vehicle traffic (1) to protected (3). We keep this distribution as-is in our proof-of-concept deployment, due to the general safety and traffic calming effect associated with protected bike infrastructure \cite{younes_traffic_2024}.

\subsection{Local attitudes towards robots}
We consider local attitudes towards robots to be a future-facing indicator that is not computable or inferrable with existing data sources.

\subsection{Weather conditions}
The weather conditions indicator needs to be calculated either in real-time or immediately before a potential deployment, due to the volatility of weather systems.

\subsection{Street lighting}
Street lighting data is deemed unavailable for our proof-of-concept deployment. In NYC, one can view a partial subset of street lights via the NYC311 request database, which details street lights that have had a physical issue, as reported by residents. However, we do not know if this includes all street lights in the city, and so do not include this feature in analysis.  

\subsection{Existence of shade}
Data to represent the existence of shade is deemed unavailable for our proof-of-concept deployment. We source data that may be apt for this feature via ShadeMap \cite{shade_map_about_2024}, but were unable to procure a parsable dataset of an area as large as NYC from the developer. 

\subsection{Pedestrian flow}
We deem data for the computation pedestrian flow out-of-scope, as discussed in \autoref{sec:discussion}.

\section{Supplemental Plots}
\begin{figure*}[h!]
    \includegraphics[width=\textwidth]{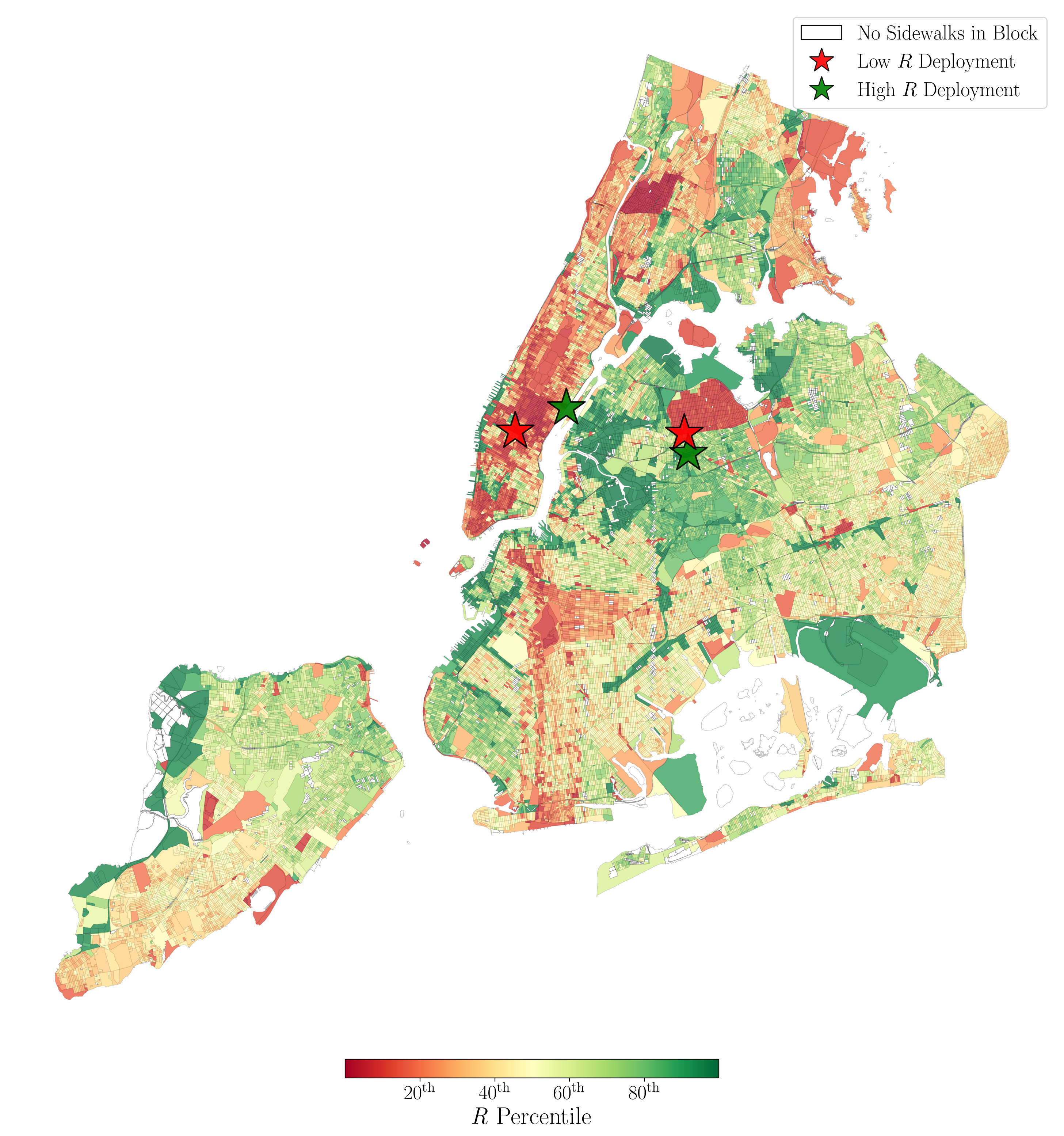}
    \caption{Robotability Score distribution in New York City, computed with the set of weights specialized for TrashBot, shown in \autoref{tab:weights}. Blocks that are colored white indicate a lack of sidewalks, or a lack of
dashcam data to estimate foundational features. Red dots indicate the locations of our two deployments in areas of lowest $R$, and green dots indicate the locations of our two deployments in areas of highest $R$.}
    \label{fig:rs-nyc-trashbot}
\end{figure*}

In this Appendix, we provide supplemental plots that reinforce or elaborate upon certain components of our analysis. We include two tables: one detailing constituent features in the traffic management systems indicator (\autoref{tab:traffic-management-features}) and another detailing constituent features in the street furniture density indicator (\autoref{tab:street-furniture-features}). Also, we include a choropleth of the proof-of-concept robotability distribution of NYC, tailored to the specific needs and requirements of TrashBot (\autoref{fig:rs-nyc-trashbot}).

\FloatBarrier

\begin{table}
    \begin{tabular}{lcl}
\toprule
\textbf{Clutter Type} & \textbf{Weight} & \textbf{OpenData Link} \\ 
\midrule
Bus Stop Shelters       & 2      & \href{https://data.cityofnewyork.us/Transportation/Bus-Stop-Shelters/qafz-7myz}{link} \\ 
Trash Can               & 0.5    & \href{https://data.cityofnewyork.us/w/8znf-7b2c/25te-f2tw?cur=CKk2SGAlT75&from=root}{link} \\ 
LinkNYC                 & 2      & \href{https://data.cityofnewyork.us/Social-Services/LinkNYC-Kiosk-Locations/s4kf-3yrf/about_data}{link} \\ 
CityBench               & 1.5    & \href{https://data.cityofnewyork.us/Transportation/City-Bench-Locations-Historical-/kuxa-tauh}{link} \\ 
Bicycle Parking Shelter & 2      & \href{https://data.cityofnewyork.us/Transportation/Bicycle-Parking-Shelters/thbt-gfu9}{link} \\ 
Bicycle Rack            & 1.5    & \href{https://data.cityofnewyork.us/Transportation/Bicycle-Parking/yh4a-g3fj}{link} \\ 
Tree                    & 0.15   & \href{https://data.cityofnewyork.us/Environment/Forestry-Tree-Points/hn5i-inap}{link} \\ 
Newsstand               & 3      & \href{https://data.cityofnewyork.us/Transportation/Newsstands/kfum-nzw3}{link} \\ 
Parking Meter           & 0.15   & \href{https://data.cityofnewyork.us/Transportation/Parking-Meters-Locations-and-Status-Map-/mvib-nh9w}{link} \\ 
Scaffolding             & 2      & \href{https://data.cityofnewyork.us/Housing-Development/NYC-Scaffold-Permits/29du-2wzn}{link} \\ 
Fire Hydrant            & 0.25   & \href{https://data.cityofnewyork.us/Environment/NYCDEP-Citywide-Hydrants/6pui-xhxz}{link} \\ 
Street Signs            & 0.05   & \href{https://data.cityofnewyork.us/Transportation/Street-Sign-Work-Orders/qt6m-xctn}{link} \\ 
\bottomrule
\end{tabular}

    \caption{Table of street furnitures used to compute the ``density of street furniture'' feature, along with the weight assigned to each type. The weight roughly approximates the spatial area that each street furniture takes up on the sidewalk.}
    \label{tab:street-furniture-features}
\end{table}

\begin{table}
    \begin{tabular}{lcl}
\toprule
\textbf{Feature Name} & \textbf{Weight} & \textbf{OpenData Link} \\ 
\midrule
Neighborhood Slow Zones       & 1      & \href{https://data.cityofnewyork.us/Transportation/VZV_Neighborhood-Slow-Zones/y4nf-25nw}{link} \\ 
Turn Traffic Calming & 1  & \href{https://data.cityofnewyork.us/Transportation/VZV_Turn-Traffic-Calming/hz4p-9f7s}{link} \\ 
Street Improvement Project (SIP) Corridors & 1      & \href{https://data.cityofnewyork.us/Transportation/VZV_Street-Improvement-Projects-SIPs-Corridor/wqhs-q6wd}{link} \\ 
SIP Intersections               & 1    & \href{https://data.cityofnewyork.us/Transportation/VZV_Street-Improvement-Projects-SIPs-intersections/79sh-heg3}{link} \\ 
Barnes Dance Intersections & 1      & \href{https://data.cityofnewyork.us/Transportation/Exclusive-Pedestrian-Signal-Barnes-Dance-Locations/8kuj-2n3u/about_data}{link} \\ 
Leading Pedestrian Intervals           & 1    & \href{https://data.cityofnewyork.us/Transportation/VZV_Leading-Pedestrian-Interval-Signals/mqt5-ctec}{link} \\ 
\bottomrule
\end{tabular}

    \caption{Table of data features used to compute the ``traffic management systems'' feature. The weight assigned to each feature is 1; they are combined summatively.}
    \label{tab:traffic-management-features}
\end{table}

\end{document}